\def\year{2022}\relax
\documentclass[letterpaper]{article} 
\usepackage{aaai22}  
\usepackage{times}  
\usepackage{helvet}  
\usepackage{courier}  
\usepackage[hyphens]{url}  
\usepackage{graphicx} 
\usepackage{natbib}  
\usepackage{caption} 
\DeclareCaptionStyle{ruled}{labelfont=normalfont,labelsep=colon,strut=off} 
\usepackage{url}
\usepackage{color}
\usepackage{multirow}
\usepackage{threeparttable}
\usepackage{booktabs}
\usepackage{amsmath}
\usepackage{amssymb}

\usepackage{algorithm}  
\makeatletter
\newif\if@restonecol
\makeatother

\usepackage{algpseudocode}
\usepackage{amsmath}

\usepackage[switch]{lineno}
\usepackage{hyperref}
\usepackage{color}

\title{{GraFormer: Graph Convolution Transformer for 3D Pose Estimation}}

\author{Weixi Zhao$^{*}$, Yunjie Tian\footnote{Contribute equally.}, Qixiang Ye, Jianbin Jiao and Weiqiang Wang\\}

\affiliations{
University of Chinese Academy of Sciences, Beijing, China \\
$\{$zhaoweixi19, tianyunjie19$\}$@mails.ucas.ac.cn, $\{$qxye, jiaojb, wqwang$\}$@ucas.ac.cn
}

\begin{document}
\maketitle

\begin{abstract}
Exploiting relations among 2D joints plays a crucial role yet remains semi-developed in 2D-to-3D pose estimation. To alleviate this issue, we propose GraFormer, a novel transformer architecture combined with graph convolution for 3D pose estimation. The proposed GraFormer comprises two repeatedly stacked core modules, GraAttention and ChebGConv block. GraAttention enables all 2D joints to interact in global receptive field without weakening the graph structure information of joints, which introduces vital features for later modules. Unlike vanilla graph convolutions that only model the apparent relationship of joints, ChebGConv block enables 2D joints to interact in the high-order sphere, which formulates their hidden implicit relations. We empirically show the superiority of GraFormer through conducting extensive experiments across popular benchmarks. Specifically, GraFormer outperforms state of the art on Human3.6M dataset while using 18$\%$ parameters. The code is available at \href{https://github.com/Graformer/GraFormer}{\color{magenta}https://github.com/Graformer/GraFormer}.
\end{abstract}

\section{Introduction}
3D human pose estimation has attracted much attention recent years in computer vision as its numerous practical applications such as action recognition ~\cite{weng2017spatio, yan2018spatial,li2019actional,jiang2021skeleton}, virtual reality~\cite{gan2019air,lu2020fmkit}, etc. The
2D-to-3D human pose estimation task takes 2D joint coordinates as inputs and outputs the 3D pose target directly, which remains a challenging problem because of the very limited knowledge contained in 2D joint coordinates. Prior works ~\cite{martinez2017simple, mehta2017monocular} have shown that the 2D kinematic structure in the coordinates is vital to learn feature representations for 3D pose estimation. However, CNN-based methods are difficult to directly process these graph-structured data.

\begin{figure}
    \centering
    \includegraphics[scale=0.27]{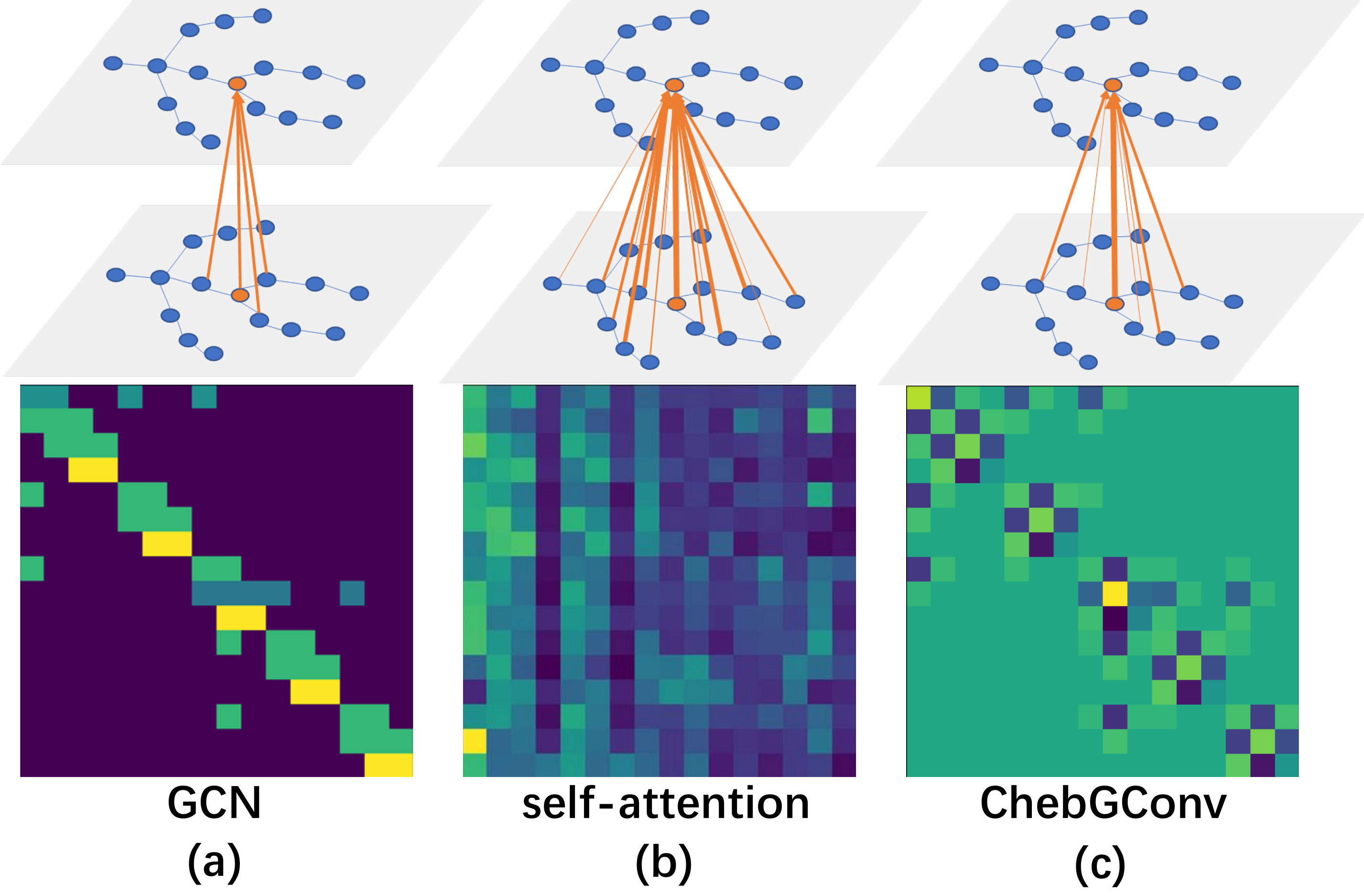}
    \caption{
    (a): the normalized adjacency matrix. The adjacency matrix activates the structurally close 2D joints and find few implicit relations. (b): the weight matrix of V in self-attention. The weight matrix is able to find the implicit relationship of 2D joints, however, which is based on the coordinate values instead of the graph structure. As a complementary, the graph Laplacian (c) from ChebGConv block can not only find many implicit relations but also remain the graph characteristics of 2D joints.}
    \label{fig:motivation} 
\end{figure}

To conquer this difficulty, recent literature ~\cite{doosti2020hope,zhao2019semantic, liu2020comprehensive, xu2021graph} try to employ graph convolution networks (GCNs) to learn the representation of these graph-structured data because of their capabilities of learning structure information.
These techniques achieve good performances but suffer from limited receptive field when learn better representations. This is because they restrict the graph convolution filter to operate only on the first-order neighbor nodes in the manner of ~\cite{kipf2016semi} as illustrated Fig.~\ref{fig:motivation} (a). 
This issue can be alleviated by stacking multiple GCN layers, while the performance may degrade due to the over-smoothing problem.
Other attempts ~\cite{zhao2019semantic} utilize non-local modules to increase the network receptive field and ~\cite{lin2021end} utilize the recent popular transformer model to capture the global vision information over the entire RGB image alleviate the receptive field problem and reach the state of the art.
The self-attention modules of transformers facilitate the interaction among all 2D joints, and thus relations of them are exploited. However, as illustrated in Fig.~\ref{fig:motivation} (b), the self-attention mechanism builds upon calculating the similarities of these joints, which actually ignores or weakens the graph structure information among these 2D joints coordinates.

To address the aforementioned problems, we propose the GraAttention module to fill in these imperfections by combining the transformer and a graph convolution filter with a learnable adjacency matrix (referred to as LAM-Gconv in short) ~\cite{kipf2016semi, doosti2020hope}. 
In particular, we replace the multiple layer perceptron (MLP) of transformers with LAM-Gconv, which facilitates the training process and boosts the performance. 
The self-attention module facilitates the interaction among 2D joints based on the coordinate values. As a complementary, the LAM-Gconv utilizes graph structure to further boost interaction.
With both superior properties of self-attention and LAM-Gconv, our GraAttention is able to learn robust features through capturing global vision information without losing the graph structure details. 

Furthermore, we claim that graph-structured data contains richer joint relations than the apparent connections. For example, the two knee joints of humans indeed exist some implicit connections, though they are not visually connected. 
We note that self-attention is able to model this kind of relationship, yet based on the similarity calculations as noted above.
Therefore, we propose to adopt the second module of GraFormer, ChebGConv block~\cite{defferrard2016convolutional} to conquer this difficulty. The proposed ChebGConv block can exchange information according to the structural relations of nodes thus attain a larger receptive field than the vanilla graph convolutions ~\cite{kipf2016semi}.
As a result, ChebGConv enables 2D joints to exploit these implicit connections via formulating the interactions in a high-order sphere. As shown in Fig.~\ref{fig:motivation} (c), we verify the effectiveness of ChebGConv with the visualization of implicit connections and ablation studies.

We demonstrate the effectiveness of our method by conducting comprehensive evaluation and ablation studies on standard 3D benchmarks. Experimental results show that our GraFormer outperforms the state of the arts on Human3.6M~\cite{ionescu2013human3} with only 0.65M parameters. In particular, we achieve 13 first-place results out of 15 categories on Human3.6M. In addition, the superiority of GraFormer is also verified on ObMan~\cite{hasson2019learning}, FHAD~\cite{garcia2018first}, and GHD~\cite{mueller2018ganerated}. The proposed method is task-independent and thus can be easily extended to other graph regression tasks.

In brief, the core contributions of our work are:
\begin{itemize}
\item We claim that the relations of 2D joints are poorly exploited for 3D pose estimation up to now, and thus in-depth study of the relationship between 2D joints will further improve its performance.
\item We propose the novel GraFormer which comprises two modules, GraAttention and ChebGConv block. Compared to prior methods, both modules aim at further exploiting relations among 2D joints for 3D pose estimation.
\item We have conducted extensive experiments on public benchmarks, which show that our GraFormer outperforms the state-of-the-art techniques for 3D pose estimation task with much fewer parameters.

\end{itemize}

\section{Related works}

\subsection{3D Pose Estimation}
 One-stage 3D pose estimation methods usually directly estimate 3D pose using image features.
 ~\cite{tekin2016direct} regresses 3D pose from a spatio-temporal volume of bounding boxes in the central frame.
 ~\cite{pavlakos2017coarse} first predicts the 3D heat map and then yields the 3D pose.
 ~\cite{mehta2017monocular} utilizes transfer learning to produce multi-modal data, which is fused to predict 3D pose.
 ~\cite{tekin2019h+} utilize 3D YOLO ~\cite{redmon2017yolo9000} model combined with the temporal information to predict the 3D pose of hand and object simultaneously.
 ~\cite{li2021exploiting} groups and predicts the joint points in a multi-tasks manner.
 ~\cite{lin2021end} takes the features extracted from the images as inputs of a transformer to predict 3D pose.

Differently, multi-stage methods first adopt CNN networks to detect 2D joint coordinates, which then are used as inputs for 3D pose estimation.
To yield 3D pose, ~\cite{chen20173d} proposes to match 2D coordinates with a 3D pose database. Based on 2D coordinates, 
 ~\cite{martinez2017simple} proposes a simple and effective network architecture using linear layers, batch normalization, dropout and ReLU activation function to regress the 3D pose. 
 ~\cite{simon2017hand} proposes a multi-view method that estimates 3D Hand pose by triangulating multiple 2D joints coordinates.
 ~\cite{hossain2018exploiting} considers 2D coordinate information as a sequence and utilizes temporal information to predict the 3D coordinates in a sequence manner. 
 
 
\begin{figure*}[htbp]
\centering
\includegraphics[scale=0.52]{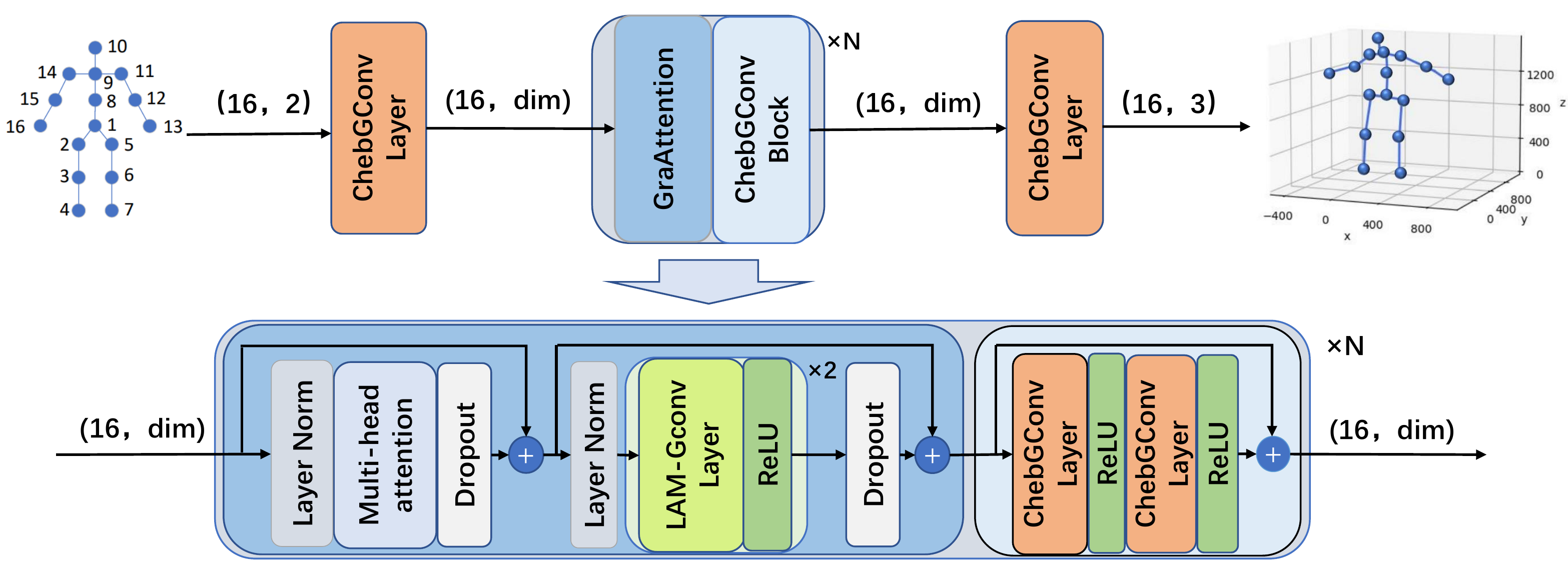}
\caption{Framework of GraFormer. The core part is the stack of GraAttention and ChebGConv block, which boosts performance for 2D-to-3D pose estimation tasks by exploiting relations among 2D joints.}
\label{fig:pipeline}
\end{figure*}

\subsection{Transformer-based Methods}
Instead of aggregating neighbor information in equal proportions according to the adjacency matrix in other works ~\cite{kipf2016semi, ge20193d},
GAT~\cite{velivckovic2017graph} proposes to use self-attention to learn the weight of each node to aggregate neighbor information.
aGCN~\cite{yang2018graph} works in the same way as GAT and learns the weights of neighbor joints through the self-attention mechanism. The difference lies in that aGCN uses the different activation functions and transformation matrix.
Although GAT and aGCN improve the effectiveness by aggregating neighbor joints, the limited receptive field remains a challenging problem. 
To attain the global receptive field, ~\cite{zhao2019semantic} utilizes the non-local layer~\cite{wang2018non} to learn the relationships between 2D joints.
~\cite{lin2021end} modifies the standard transformer encoder and adjusts the dimension of encoder layers. However, such interaction ignores the graph structure as noted above.
Our method increases the receptive field through self-attention mechanism and rediscovers graph structure by GCNs to effectively improve the performance for 3D pose estimation.

\subsection{GCN-based methods}

Recently, some works in the 3D pose estimation task have achieved state-of-the-art results by using graph convolutional networks, such as ~\cite{ge20193d, zhao2019semantic, doosti2020hope, liu2020comprehensive, xu2021graph}.
~\cite{ge20193d} uses a stacked hourglass network~\cite{newell2016stacked} to extract features from images, which are later reshaped to the graph structure.
The 3D mesh is predicted using the extracted features by graph convolution, which then is used to predict the 3D pose.
~\cite{zhao2019semantic} proposes semantic graph convolution, which learns the weights among neighbor joints. Non-local modules are used to enhance interaction among 2D joints.
~\cite{doosti2020hope} proposes modified graph pooling and unpooling operations, which make the up-sampling and down-sampling procedures trainable for graph-structured data.
~\cite{xu2021graph} proposes graph hourglass network and adopts SE Block~\cite{hu2018squeeze} to fuse features extracted from different layers of graph hourglass network.
In this paper, we use graph convolution operations combined with transformer to solve the 3D pose estimation problem.

\section{Method}

The proposed framework is shown in Fig.~\ref{fig:pipeline}. Our method takes 2D joint coordinates as inputs and predicts the target 3D poses. GraFormer is a combination of transformer and graph convolution, which much further boosts the interaction of 2D joints to exploit the relations among them compared to prior works  ~\cite{zhao2019semantic, doosti2020hope, liu2020comprehensive, xu2021graph}. 
GraFormer is formed by stacking GraAttentions and ChebGConv blocks, which are described in detail as below.

\subsection{Preliminaries}

Graph convolution operation and multi-head self-attention are the fundamental building elements of GraFormer.
GCNs have the ability to handle graph-structured data because of the specific design described below.
Assume that $X^l\in R^{j\times D_l}$ is the input of the $l$-th layer of GCN, which contains $j$ joints and $X_{i}^{l}\in R^{D_l}$. $D^l$ denotes the input dimension. The initial input of GraFormer, $X^0\in R^{j\times 2}$, is the 2D joint coordinates of human body or hand. The output of a GCN layer can be formulated as:

\begin{equation}
  X^{l+1}=\sigma \left( \tilde{D}^{-\frac{1}{2}}\tilde{A}\tilde{D}^{-\frac{1}{2}}X^l\Theta \right) 
\end{equation}
where $\sigma $ is the activation function, $\Theta \in R^{D_l\times D_{l+1}}$ denotes a learnable weight matrix, $A\in R^{j\times j}$ is the adjacency matrix, $\widetilde{A}=A+I$ and $\widetilde{D}$ is the diagonal node degree matrix. We make $\widetilde{A}$ learnable for LAM-Gconv in GraAttention.

The gains achieved by transformer-based methods ~\cite{lin2021end} mainly come from the global vision view, which benefits from the self-attention mechanism as described below. We use the same notation of $X^l\in R^{j\times D_l}$ as input, which is put into a MLP layer to produce a feature $Y^l$ with the dimension of $j\times D$, and then be cut into 3 portions, referred to as $Q^l$, $K^l$ and $V^l$ respectively. Thereafter, $X^{l+1}$ is attained by:
\begin{equation}
  X^{l+1}=\sigma(\frac{Q^l\cdot{K^l}^{T}}{\sqrt{D}})V^l
\end{equation}


\subsection{GraAttention}
From this section, we start to introduce two GraFormer core modules, GraAttention and ChebGConv block.
As illustrated in Fig.~\ref{fig:pipeline}, 
the 16 2D joint coordinates are first pre-processed by a ChebGConv layer (which will be introduced below) followed by the stack of GraAttention and ChebGConv block. Thereafter, 16 feature vectors with the dimension of $dim$ (referred to as feature dimension) are yielded, which then are put into GraAttention. Before the multi-head self-attention block in GraAttention, the input feature vectors are first normalized by layer norm (LN), which is commonly used in transformer models ~\cite{vaswani2017attention}.

The multi-head self-attention block is inherited from the transformer encoder layers~\cite{vaswani2017attention}. Different from other applications of transformer on 3D pose estimation like ~\cite{lin2021end}, we remove the MLP layer from the standard transformer. Because we observe that the MLP impedes the learning procedure as shown in experiments. Then, the self-attention output is regularized by a dropout layer ~\cite{srivastava2014dropout}.
Each element of the output of the multi-head attention block contains all 2D joint information because the global interaction on the graph consists of all the 2D joints, such that the non-local relationships can be exploited.

Next, the output is normalized by the LN layer followed by GCN layers and a ReLU activation. Different from vanilla graph convolution operation, we make the adjacency matrix to be learnable thus that the GCN layer could be more flexible to learn graph-structured data. The GCN layer is referred to as LAM-Gconv as noted above. Next, a dropout layer is followed.
GraAttention is a combination of multi-head self-attention block and GCN layer, and both blocks include a shortcut connection as shown in Fig.~\ref{fig:pipeline}.

\subsection{ChebGConv block}
The second graph convolution operation we used is Chebyshev graph convolution~\cite{defferrard2016convolutional}, referred to as ChebGConv in short. Compared with traditional GCN layers, ChebGConv is more powerful to handle with graph-structured data, which can be formulated as below.
We first introduce its calculation formula.
The normalized graph Laplacian is computed as:
\begin{equation}
    L=\,\,I-\tilde{D}^{-\frac{1}{2}}A\tilde{D}^{-\frac{1}{2}}
\end{equation}
The formula of Chebyshev graph convolution is:
\begin{equation}
X^{l+1}=\sum_{k=0}^{K-1}{T_k\left( \tilde{L} \right)}X^l\theta _k 
\end{equation}
where $T_k\left( x \right) =2xT_{k-1}\left( x \right) -T_{k-2}\left( x \right) $ denotes the Chebyshev
polynomial of degree $k$, $T_0=1,T_1=x$, and $\tilde{L}\in R^{j\,\,\times j}$ denotes the
rescaled Laplacian, $\tilde{L}=2L/\lambda _{\max}-I$, $\lambda _{\max}$ is the maximum eigenvalue of $L$.
$\theta _k\in R^{D_l\times D_{l+1}}$ denotes the trainable parameters in the graph convolutional layer. Since the convolution kernel is a K-order polynomial graph Laplacian, ChebGConv block is able to fuse information among the K top neighbors of a joint, which brings a larger receptive field.
Coupled with the characteristic of graph convolution filter, ChebGConv block boosts the performance as verified in experiments.

Even though the more expensive computation than the traditional GCN layer, ChebGConv block complexity increases slightly because the graph is simple with only 16 joints. 

\subsection{Training}
To train GraFormer, we apply a loss function to the final output to minimize the error between the 3D predictions and ground truth. Given dataset $S=\left\{ J_{i}^{2d},J_{i}^{3d} \right\} _{i=1}^{N}$, where $J_{i}^{2d}\in R^{j\times 2}$ is 2D joints of the human body or hand, $j$ is set to 16 for the human datasets and 21 for the hand datasets.
$J_{i}^{3d}\in R^{j\times 3}$ is the 3D ground truth coordinates. $N$ denotes the total number of training samples.

We use the mean squared errors (MSE) to minimize the error between the 3D predictions and ground truth coordinates.
\begin{equation}
    L=\frac{1}{N}\sum_{i=1}^N{\left( \left\| \tilde{J}_{i}^{3d}-J_{i}^{3d} \right\| ^2 \right)}
\end{equation}
where $\tilde{J}_{i}^{3d}\in R^{j\times 3}$ denotes the predicted 3D coordinates. We calculate all 3D coordinate errors in millimeters.

\section{Experiments}

In this section, we first introduce the experimental details and training settings. Next, we compare GraFormer with other state-of-the-art methods and analyze the results. Finally, we conduct ablation studies to verify the effectiveness of GraFormer.

\subsection{Experimental Details}
\paragraph{Dataset}
We use 3 popular hand datasets, including ObMan ~\cite{hasson2019learning}, FHAD ~\cite{garcia2018first}, GHD~\cite{mueller2018ganerated}, and 1 human pose dataset Human3.6M ~\cite{ionescu2013human3} to evaluate our GraFormer. 
ObMan is a large synthetic dataset of hand-object interaction scenarios. The hands are generated from MANO~\cite{romero2017embodied} and the objects are selected from the Shapenet~\cite{chang2015shapenet} dataset. The ObMan dataset contains 141K training frames and 6K evaluation frames. Each frame contains an RGB image, a depth image, a 3D mesh of the hand and object, and 3D coordinates for the hand.

FHAD ~\cite{garcia2018first} contains videos of manipulating different objects from the first-person perspective. There are a total of 21,501 frames of images, where 11019 frames are used for training and 10482 frames for testing.

There are 143,449 frames without objects, and 188,050 frames with objects in the GHD~\cite{mueller2018ganerated} dataset. The dataset is divided into groups and each group consists of 1024 frames. 
There are a total of 141 groups without objects and we use the first 130 groups for training and the last 11 groups for testing.

Human3.6M ~\cite{ionescu2013human3} is the most widely used dataset in the 3D human pose estimation task. It provides 3.6M accurate 3D poses captured by the MoCap system in the indoor environment. 
It contains 15 actions performed by seven actors from four cameras.
There are two common evaluation protocols by splitting training and testing set in prior methods~\cite{martinez2017simple,zhao2019semantic, liu2020comprehensive,xu2021graph}.  
The first protocol uses subjects S1, S5, S6, S7 and S8 for training, and S9 and S11 for testing. Errors are calculated after the ground truth and predictions are aligned with the root joints.
The second protocol uses subjects S1, S5, S6, S7, S8 and S9 for training, and S11 for testing. 
We conduct experiments using the first protocol. In this way, there are 1,559,752 frames for training, and 543,344 frames for testing.

\begin{table*}[htb]
	\centering
	\begin{threeparttable}
    	\resizebox{\textwidth}{30mm}{
         \begin{tabular}{l|ccccccccccccccc|c}
          \toprule 
	       Methods & Direct. & Discuss & Eating & Greet & Phone & Photo & Pose & Purch. & Sitting & SittingD. & Smoke & Wait & WalkD. & Walk & WalkT. & Avg. \\
	        \midrule 
	            \cite{ionescu2013human3} PAMI & 132.7 & 183.6 & 132.3 & 164.4 & 162.1 & 205.9 & 150.6 & 171.3 & 151.6 & 243.0 & 162.1 & 170.7 & 177.1 & 96.6 & 127.9 & 162.1\\
            	\cite{tekin2016direct} CVPR & 102.4 & 147.2 & 88.8 & 125.3 & 118.0 & 182.7 & 112.4 & 129.2 & 138.9 & 224.9 & 118.4 & 138.8 & 126.3 & 55.1 & 65.8 & 125.0 \\
                \cite{zhou2016deep} CVPR & 87.4 & 109.3 & 87.1 & 103.2 & 116.2 & 143.3 & 106.9 & 99.8 & 124.5 & 199.2 & 107.4 & 118.1 & 114.2 & 79.4 & 97.7 & 113.0 \\
            	\cite{du2016marker} ECCV & 85.1 & 112.7 & 104.9 & 122.1 & 139.1 & 135.9 & 105.9 & 166.2 & 117.5 & 226.9 & 120.0 & 117.7 & 137.4 & 99.3 & 106.5 & 126.5\\
            	\cite{chen20173d} CVPR     & 89.9 & 97.6 & 89.9 & 107.9 & 107.3 & 139.2 & 93.6 & 136.0 & 133.1 & 240.1 & 106.6 & 106.2 & 87.0 & 114.0 & 90.5 & 114.1\\
            	\cite{pavlakos2017coarse} CVPR & 67.4 & 71.9 & 66.7 & 69.1 & 72.0 & 77.0 & 65.0 & 68.3 & 83.7 & 96.5 & 71.7 & 65.8 & 74.9 & 59.1 & 63.2 & 71.9\\
            	\cite{mehta2017monocular} 3DV &52.6 & 64.1 & 55.2 & 62.2 & 71.6 & 79.5 & 52.8 & 68.6 & 91.8 & 118.4 & 65.7 & 63.5 & 49.4 & 76.4 & 53.5 & 68.6\\
            	\cite{zhou2017towards} ICCV & 54.8 & 60.7 & 58.2 & 71.4 & 62.0 & 65.5 & 53.8 & 55.6 & 75.2 & 111.6 & 64.1 & 66.0 & 51.4 & 63.2 & 55.3 & 64.9\\
            	\cite{martinez2017simple} ICCV & 51.8 & 56.2 & 58.1 & 59.0 & 69.5 & 78.4 & 55.2 & 58.1 & 74.0 & 94.6 & 62.3 & 59.1 & 65.1 & 49.5 & 52.4 & 62.9\\
            	\cite{sun2017compositional} ICCV  & 52.8 & 54.8 & 54.2 & 54.3 & 61.8 & 53.1 & 53.6 & 71.7 & 86.7 & 61.5 & 67.2 & 53.4 & 47.1 & 61.6 & 53.4 & 59.1\\
            	\cite{fang2018learning} AAAI & 50.1 & 54.3 & 57.0 & 57.1 & 66.6 & 73.3 & 53.4 & 55.7 & 72.8 & 88.6 & 60.3 & 57.7 & 62.7 & 47.5 & 50.6 & 60.4\\
                \cite{zhao2019semantic} CVPR & 48.2 & 60.8 & 51.8 & 64.0 & 64.6 & 53.6 & 51.1 & 67.4 & 88.7 & 57.7 & 73.2 & 65.6 & 48.9 & 64.8 & 51.9 & 60.8\\
            \midrule 
            GraFormer (HG) & 49.3 & 53.9 & 54.1 & 55.0 & 63.0 & 69.8 & 51.1 & 53.3 & 69.4 & 90.0 & 58.0 & 55.2 & 60.3 & 47.4 & 50.6 & 58.7\\
            \midrule 
                \cite{martinez2017simple} (GT)& 37.7 & 44.4 & 40.3 & 42.1 & 48.2 & 54.9 & 44.4 & 42.1 & 54.6 & 58.0 & 45.1 & 46.4 & 47.6 & 36.4 & 40.4 & 45.5 \\
                
                \cite{hossain2018exploiting} (GT) & 35.2 & 40.8 & 37.2 & 37.4 & 43.2 & 44.0 & 38.9 & 35.6 & 42.3 & 44.6 & 39.7 & 39.7 & 40.2 & 32.8 & 35.5 & 39.2 \\
                
                \cite{zhao2019semantic} (GT) & 37.8 & 49.4 & 37.6 & 40.9 & 45.1 & \textbf{41.4} & 40.1 & 48.3 & 50.1 & \textbf{42.2} & 53.5 & 44.3 & 40.5 & 47.3 & 39.0 & 43.8 \\
                \cite{liu2020comprehensive} (GT) & 36.8 & 40.3 & 33.0 & 36.3 & 37.5 & 45.0 & 39.7 & 34.9 & 40.3 & 47.7 & 37.4 & 38.5 & 38.6 & 29.6 & 32.0 & 37.8 \\
                \cite{xu2021graph} (GT) &  35.8 & 38.1 & 31.0 & 35.3 & 35.8 & 43.2 & 37.3 & 31.7 & 38.4 & 45.5 & 35.4 & 36.7 & 36.8 & 27.9 & 30.7 & 35.8\\
            \midrule 
               GraFormer (GT) & \textbf{32.0} & \textbf{38.0} & \textbf{30.4} & \textbf{34.4} & \textbf{34.7} & 43.3 & \textbf{35.2} & \textbf{31.4} & \textbf{38.0} & 46.2 & \textbf{34.2} & \textbf{35.7} & \textbf{36.1} & \textbf{27.4} & \textbf{30.6} & \textbf{35.2}\\
	        \bottomrule
    	\end{tabular}
    	}
	\end{threeparttable}
	\caption{MPJPE (mm) results on Human3.6M. This table is split into 2 groups. The inputs for the top group methods are images and the inputs for the bottom group are ground truth of 2D joints.  We note that GraFormer achieves the state-of-the-art results with only \textbf{0.65M} parameters compared to ~\cite{xu2021graph} with \textbf{3.70M} parameters.}
	\label{tab:human36m}
\end{table*}

\begin{table}[htb]
	\centering
	\begin{threeparttable}
	\scalebox{0.9}{
        \begin{tabular}{lcccc}
          \toprule 
	       Methods & ObMan & FHAD & GHD \\
	        \midrule 
	           Linear \cite{martinez2017simple} & 23.64 &26.15 & 39.25\\
            	Graph U-Net\cite{doosti2020hope} & 7.63 & 13.82 & 8.45 \\
            	GraFormer(ours) &  3.29 & 11.68 & 4.25\\
	        \bottomrule
    	\end{tabular}
    	}
	\end{threeparttable}
	\caption{MPJPE (mm) results compared with Linear model and  Graph U-Net on three hand datasets, ObMan, FHAD and GHD.}
	\label{tab:hand_result}
\end{table}

\paragraph{Evaluation Metric}
We follow the same evaluation metric as in ~\cite{zhao2019semantic}.
The evaluation metric is the Mean Per Joint Position Error (MPJPE) in millimeters, which is calculated between the ground truth and the predicted 3D coordinates across all cameras and joints after aligning the predefined root joints (the pelvis joint).

\paragraph{Training Settings}

For the three hand datasets ObMan, FHAD and GHD, we directly take 2D image coordinates and 3D camera coordinates as the inputs and ground truth.
The 2D coordinates and 3D ground truth provided by ObMan can be used by simply converting the ground truth from meter to millimeter. The 3D ground truth provided by FHAD are world coordinates, and we use the extrinsic matrix to calculate the corresponding camera coordinates. 
The 3D ground truth of GHD are already the camera coordinate system, and the 2D coordinates need to be cropped, scaled and restored.
We use the hand coordinates as input but object coordinates for all hand datasets.
For Human3.6M, because of the multiple camera views, it needs to be normalized according to ~\cite{zhao2019semantic} before training and evaluation.


In our experiment, we set the number of N in Fig.~\ref{fig:pipeline} to 5 and adopt 4 heads for self-attention. Different from the feature dimension value of 64 or 128 in prior works ~\cite{zhao2019semantic, xu2021graph}, we set the middle feature dimension of the model to 96 for GraFormer with a dropout rate of 0.25.
We adopt Adam ~\cite{kingma2014adam} optimizer for optimization with an initial learning rate of 0.001 and mini-batches of 64. For Human3.6M, we multiply the learning rate by 0.9 every 75000 steps. 
For hand datasets, the learning rate decays by 0.9 every 30 epochs. 
We train GraFormer for 50 epochs on Human3.6M, 900 epochs on Obman and GHD and 3000 epochs on FHAD.

\subsection{Performance and Comparison}

In this section, we evaluate our GraFormer on several popular datasets and analyze the performances compared with other state-of-the-art methods.
\paragraph{Performance on Hand Datasets}{
We first verify GraFormer on hand datasets via comparing with Graph U-Net ~\cite{doosti2020hope} and Linear model ~\cite{martinez2017simple} since all of these methods regress 3D pose results by taking 2D ground truth coordinates as inputs.
The results on 3 hand datasets are shown in Tab.~\ref{tab:hand_result}. One can see that GraFormer achieves the best performance across all these 3 hand datasets. In particular, GraFormer surpasses Graph U-Net by a large margin of $4.34$, $2.14$ and $4.2$ on ObMan, FHAD and GHD respectively. We note that small datasets are not friendly to self-attention, even so, GraFormer still beats Graph U-Net on the severely small dataset FHAD.}

\begin{figure*}[htbp]
\centering
\includegraphics[scale=0.45]{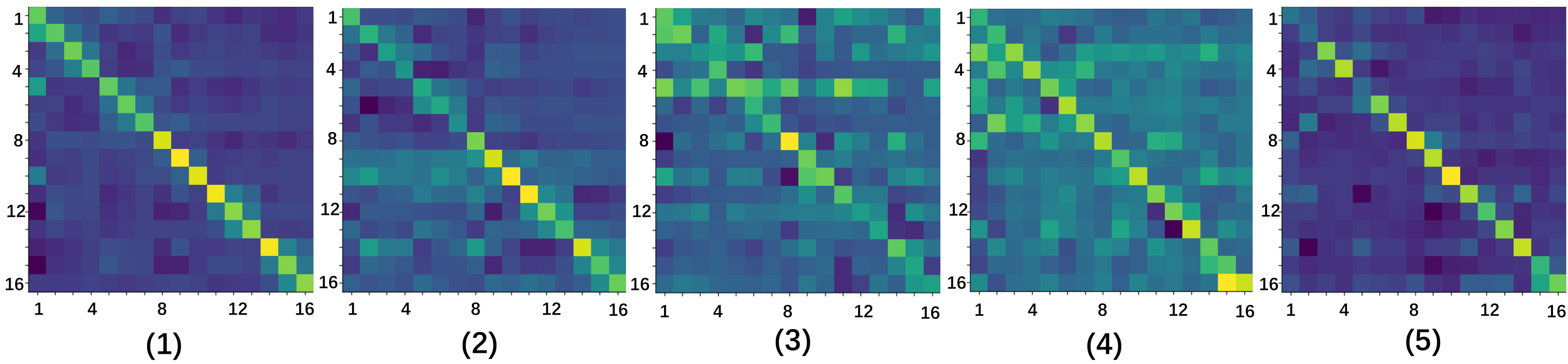}
\caption{
Visualization of the learned adjacency matrixes of different LAM-Gconv layers in the GraAttention module. It can be observed that the adjacency matrixes of the first and last layers mainly aggregate information among their near neighbor joints. While the middle 3 layers aggregate long-range information.}
\label{fig:adj_matrix_viz}
\end{figure*}

\paragraph{Performance on Human Pose Dataset}{
Prior work results on Human3.6M can be divided into two groups as shown in Tab.~\ref{tab:human36m}. The top group methods take the images as inputs and then yield the 3D poses using the learned features. For our method, we adopt 2D coordinate detection results provided by ~\cite{zhao2019semantic}, which are detected by stacked hourglass network~\cite{newell2016stacked}. In the top group, we only compare our method with the methods before \cite{zhao2019semantic}. Because the new methods use more advanced 2D detectors, which can provide more accurate 2D coordinates but their codes are not public available. In the bottom group, we compare with the most advanced methods using the same 2D inputs. The results show that GraFormer surpasses all the methods, which evaluates the superiority of our method.

The bottom group methods take 2D ground truth as inputs to predict 3D pose coordinates directly. 
Compared with the previous methods, GraFormer achieves the best performance which evaluates the effectiveness of our method. In particular, GraFormer obviously improves scores in direction, eating, greet, phone, pose, smoke, and wait with only 0.65M parameters (18\% of ~\cite{xu2021graph}). 
Interestingly, we find that these actions have large magnitude motions, which implies longer distances among 2D joints than the actions of Photo and SittingD. This means GraFormer has a more powerful capability to capture information of 2D graph joints with larger magnitude action motions. 

}

\begin{table}[htb]
	\centering
	\begin{threeparttable}
	\scalebox{0.9}{
        \begin{tabular}{lcccc}
          \toprule 
	       Methods & params & MPJPE(mm)  \\
	        \midrule 
	            GAT\cite{velivckovic2017graph} & 0.16M & 82.9  \\
            	ST-GCN\cite{yan2018spatial} & 0.27M & 57.4  \\
                FC\cite{martinez2017simple} & 4.29M & 45.5  \\
                SemGCN\cite{zhao2019semantic} & 0.43M & 43.8 \\
                Pre-agg\cite{liu2020comprehensive} & 4.22M & 37.8 \\
                GraphSH\cite{xu2021graph} & 3.70M & 35.8 \\
            \midrule
                GraFormer-small (ours) & 0.12M & 38.9\\ 
            	GraFormer (ours) & 0.65M & 35.2 \\
	        \bottomrule
    	\end{tabular}
        }
	\end{threeparttable}
	\caption{Results on Human3.6M dataset under different parameter configurations.}
	\label{tab:params}
\end{table}

\subsection{Ablation Study}
\paragraph{Discussion on model parameters}{ 
We start our ablation experiments by comparing GraFormer on different parameter configurations with other methods on the Human3.6M dataset, Tab.~\ref{tab:params}. 
We report the results of our models of two configurations to prove that our method can achieve better results with fewer parameters than other methods. 
GraFormer-small has only 2 layers, and the feature dimension is 64 with a dropout rate of 0.1. GraFormer sets the N to 5, and the feature dimension is set to 96 with a dropout rate of 0.25. 
Our GraFormer achieves better results with even much fewer parameters than GraphSH, etc.
Our lightweight version, GraFormer-small, with only 72$\%$ fewer parameters than SemGCN, beats SemGCN by $4.9$. 
}

\begin{table}[htb]
	\centering
	\begin{threeparttable}
        \begin{tabular}{lccc}
          \toprule 
	       Models & Human3.6M & ObMan & FHAD\\
	        \midrule 
	            model-T  & 51.76 & 15.54 & 20.14 \\
	            model-C &  47.81 & 8.51 &  16.30 \\
	            model-M &  42.19 & 5.02 & 13.52 \\
	            model-AT & 37.78 & 7.29 & 14.39 \\
	            model-AM & 42.44 & 3.46 & 13.49 \\
            	GraFormer & 35.17 & 3.29 & 11.68 \\
	        \bottomrule
    	\end{tabular}
	\end{threeparttable}
	\caption{MPJPE (mm) results on Human3.6M, ObMan and FHAD by removing or replacing GraFormer's modules.}
	\label{tab:ablation}
\end{table}

\paragraph{Effects of GraFormer Modules}{

Next, we test GraFormer modules on Human3.6M ~\cite{ionescu2013human3} and ObMan ~\cite{hasson2019learning}.
We design 5 models by removing or replacing GraFormer's modules to test the effects of our method. We keep all parameters the same for the 5 models if not particularly indicated. 
Specifically, Model-T is formed through replacing the stack of GraAttention and ChebGConv block by transformer encoder.
Model-C removes GraAttention from GraFormer.
Model-M replaces GraAttention with self-attention.
Model-AT removes the ChebGconv block from GraFormer. And model-AM reserves MLP compared to GraFormer.
The results are shown in Table ~\ref{tab:ablation}.

\begin{figure}[htbp]
\centering
\includegraphics[scale=0.25]{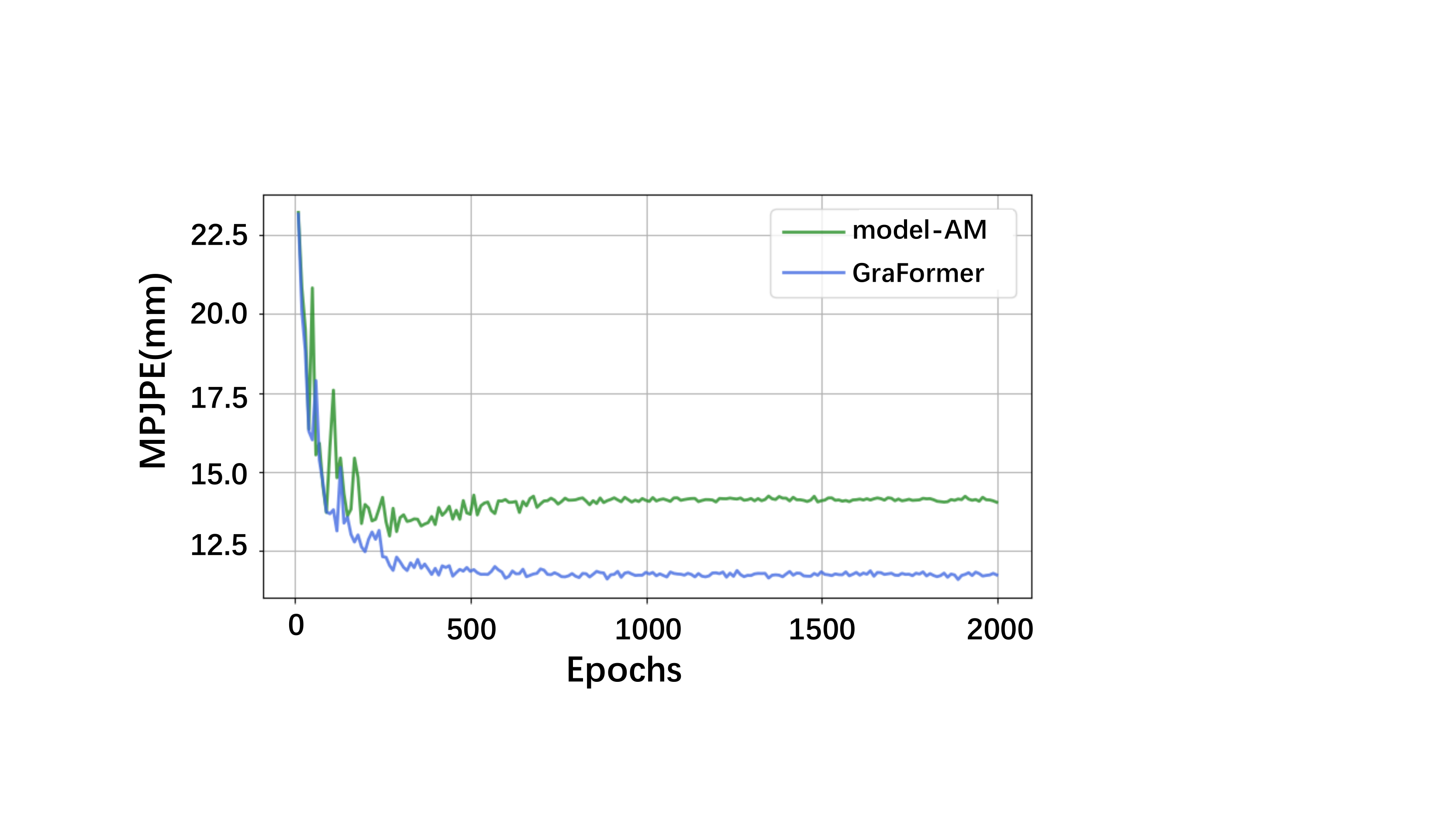}
\caption{Test errors of model-AM and GraFormer on FHAD dataset.}
\label{fig:model_am}
\end{figure}

\begin{figure*}[htbp]
\centering
\includegraphics[scale=0.53]{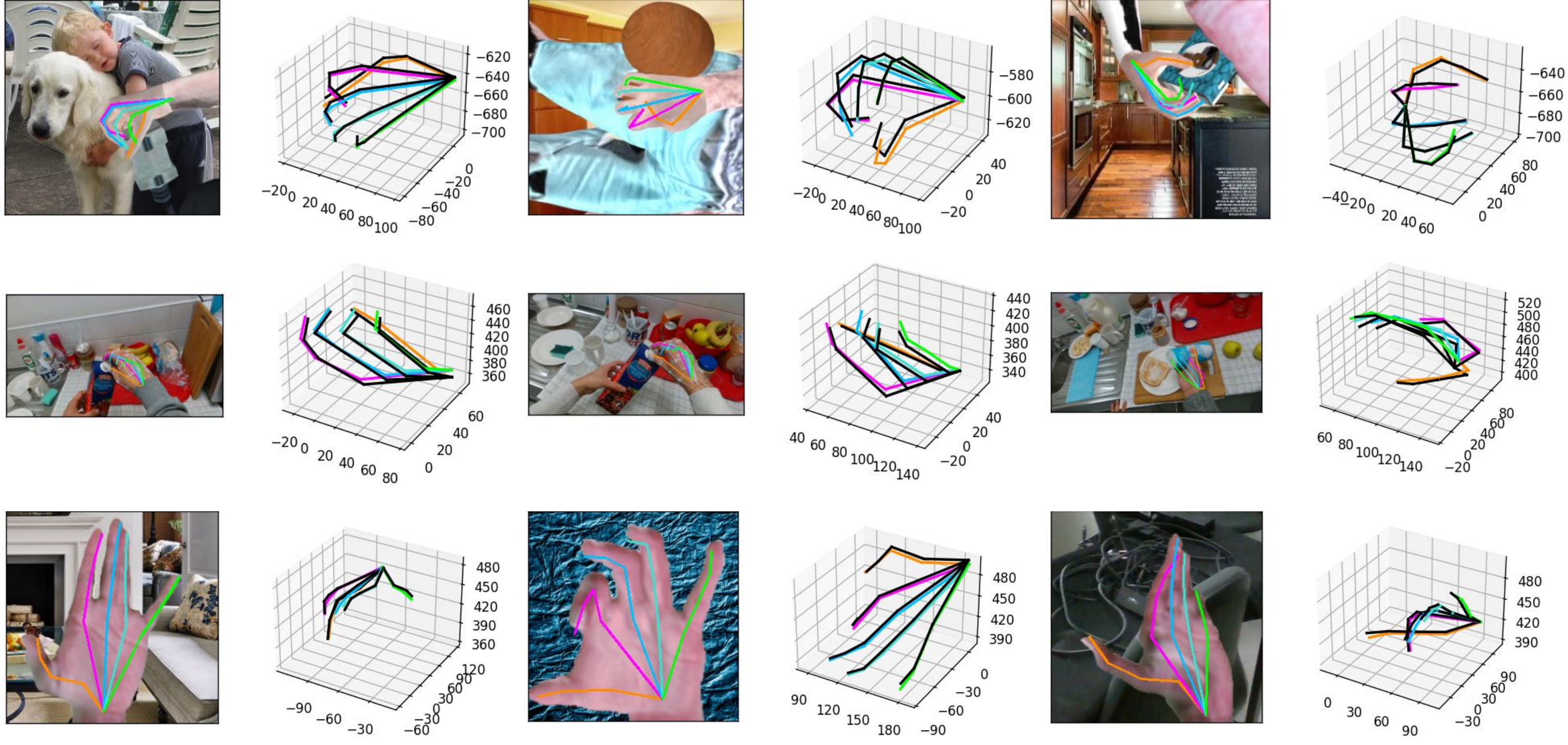}
\caption{Skeleton results predicted by GraFormer on ObMan~\cite{hasson2019learning} (top row), FHAD~\cite{garcia2018first} (middle row) and GHD~\cite{mueller2018ganerated} (bottom row).}
\label{fig:result_image}
\end{figure*}

From the results of model-T, we can find that the transformer is poor for 2D-to-3D pose estimation. This is because transformer ignores graph structure information. 
The results of model-C and model-M are worse than GraFromer, which prove that GraAttention is necessary and more effective than self-attention.
The loss of ChebGConv block in model-AT brings worse performance, which illustrates the effectiveness of ChebGConv block.
The performance also degrades when plugging the MLP layer after self-attention in GraAttention, which verifies that the MLP layer impedes the learning of 3D poses actually. In Fig.~\ref{fig:model_am}, we show the test errors of model-AM and GraFormer. We find that the test error of model-AM is hindered at about 250 epochs while test error of GraFomer still decreases until below 12. This reconfirms that the MLP layer can be a deterrent on 2D-to-3D pose estimation task.
}

\begin{figure}
    \centering
    \includegraphics[scale=0.25]{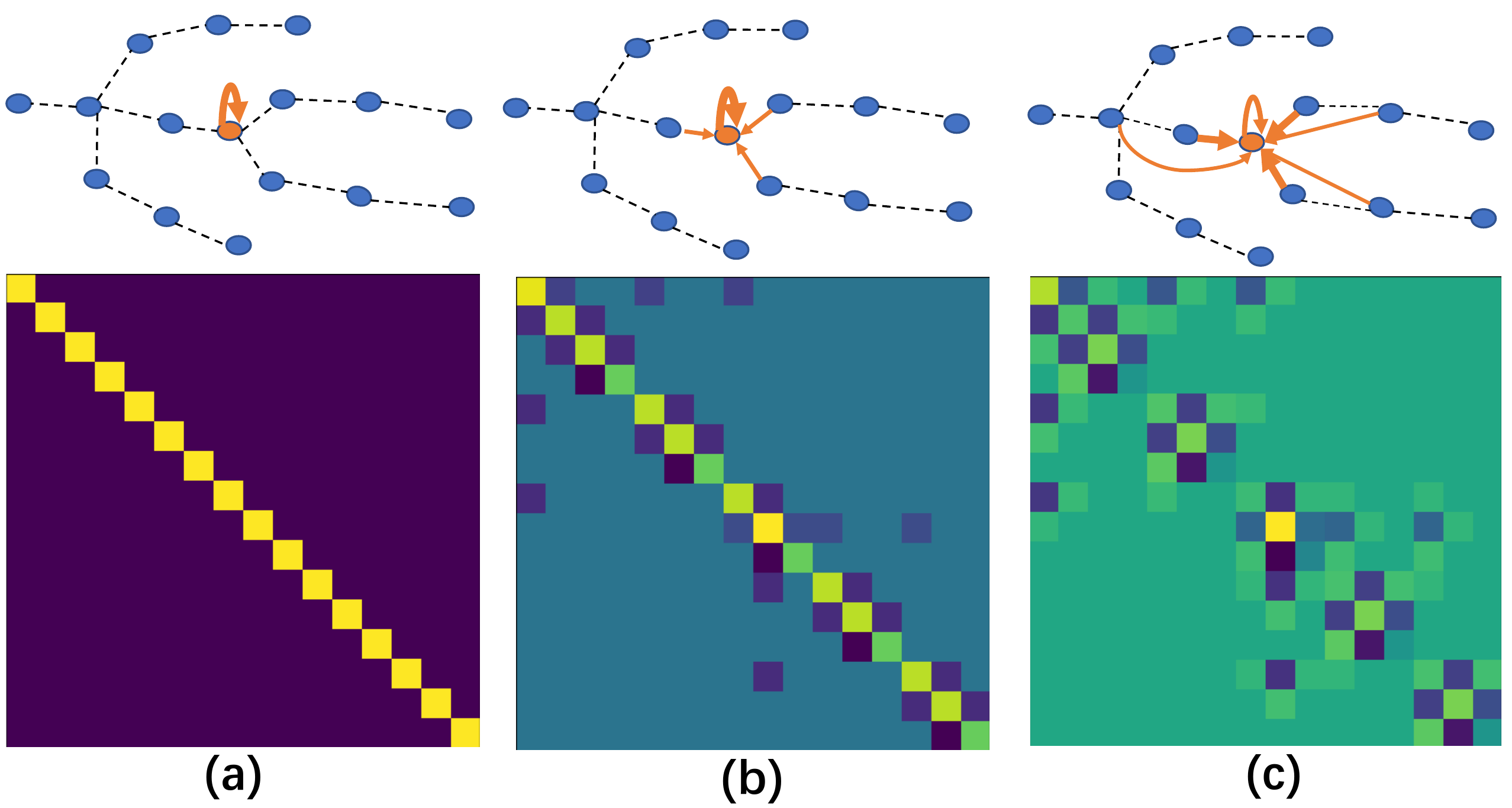}
    \caption{Visualization of graph Laplacian of ChebGConv.}
    \label{fig:cheb_viz}  
\end{figure}

\paragraph{Visualization}{
Fig.~\ref{fig:adj_matrix_viz} shows the visualization results of learned adjacency matrixes of LAM-Gconv layers of GraAttention. 
In the first sub-figure (1), the color of some 3$\times$3 regions is obviously brighter than other regions, which indicates that interaction among these joints takes greater weights and these joints are closely connected.
Interestingly, we note that joints 2-4, 5-7, 11-13 and 14-16 are four limbs of the human, which are activated in $3\times3$ regions. Which implies that the relations of joints on a limb are strongly connected, and our GraFormer is able to find these relations effectively.
The regions of sub-figures from (2) to (4) become larger, which shows that GraFormer finds long-range relationships in these layers.
The sub-figure (5) illustrates that the interaction regions become much smaller, which implies that joints mainly retain their own information, and a little information interacts among joints. 

In Figure~\ref{fig:result_image}, we show the predicted 3D hand results on ObMan~\cite{hasson2019learning} (top row), FHAD~\cite{garcia2018first} (middle row) and GHD~\cite{mueller2018ganerated} (bottom row). 
The images in columns 1, 3 and 5 are skeleton figures drawn using 2D ground truth.
In columns 2, 4 and 6, the colored skeletons are drawn using the 3D predictions, and the black skeletons are drawn using the 3D ground truth.
We can find that our method is able to estimate the 3D poses accurately using 2D coordinates. It shows that our method could effectively learns 3D poses by exploiting the relationship among 2D joints.

Figure~\ref{fig:cheb_viz} is a visualization of graph Laplacian of different orders of Chebyshev graph convolution. The top row shows the schematic diagrams of joint information aggregation according to the bottom row. 
The width of the line implies the weights between 2D joints. The bottom row shows the visualization of the corresponding Laplacian matrix with 0-order (a), 1-order (b) and 2-order (c) respectively. It is easy to find that bigger orders matrix activates more 2D joints, which implies that bigger orders of Laplacian matrix have the capability to find more implicit relations.
}

\section{Conclusions}
2D-to-3D pose estimation task takes graph-structured 2D joint coordinates as inputs. We claim that 2D joints relations remain semi-developed and the performance can be further boosted by exploiting these relations.
We propose a novel model combined graph convolution and transformer, called GraFormer, for 3D pose estimation, which aims at better exploiting relations among graph-structured 2D joints.
Extensive experiment results across popular benchmarks show that our method achieves state-of-the-art performances compared with previous works using much fewer parameters.

%

\bibliography{aaai22.bib}

\begin{thebibliography}{43}
\providecommand{\natexlab}[1]{#1}

\bibitem[{Chang et~al.(2015)Chang, Funkhouser, Guibas, Hanrahan, Huang, Li,
  Savarese, Savva, Song, Su et~al.}]{chang2015shapenet}
Chang, A.~X.; Funkhouser, T.; Guibas, L.; Hanrahan, P.; Huang, Q.; Li, Z.;
  Savarese, S.; Savva, M.; Song, S.; Su, H.; et~al. 2015.
\newblock Shapenet: An information-rich 3d model repository.
\newblock \emph{arXiv preprint arXiv:1512.03012}.

\bibitem[{Chen and Ramanan(2017)}]{chen20173d}
Chen, C.-H.; and Ramanan, D. 2017.
\newblock 3d human pose estimation= 2d pose estimation+ matching.
\newblock In \emph{IEEE CVPR}, 7035--7043.

\bibitem[{Defferrard, Bresson, and
  Vandergheynst(2016)}]{defferrard2016convolutional}
Defferrard, M.; Bresson, X.; and Vandergheynst, P. 2016.
\newblock Convolutional neural networks on graphs with fast localized spectral
  filtering.
\newblock \emph{arXiv preprint arXiv:1606.09375}.

\bibitem[{Doosti et~al.(2020)Doosti, Naha, Mirbagheri, and
  Crandall}]{doosti2020hope}
Doosti, B.; Naha, S.; Mirbagheri, M.; and Crandall, D.~J. 2020.
\newblock Hope-net: A graph-based model for hand-object pose estimation.
\newblock In \emph{IEEE CVPR}, 6608--6617.

\bibitem[{Du et~al.(2016)Du, Wong, Liu, Han, Gui, Wang, Kankanhalli, and
  Geng}]{du2016marker}
Du, Y.; Wong, Y.; Liu, Y.; Han, F.; Gui, Y.; Wang, Z.; Kankanhalli, M.; and
  Geng, W. 2016.
\newblock Marker-less 3d human motion capture with monocular image sequence and
  height-maps.
\newblock In \emph{ECCV}, 20--36.

\bibitem[{Fang et~al.(2018)Fang, Xu, Wang, Liu, and Zhu}]{fang2018learning}
Fang, H.-S.; Xu, Y.; Wang, W.; Liu, X.; and Zhu, S.-C. 2018.
\newblock Learning pose grammar to encode human body configuration for 3d pose
  estimation.
\newblock In \emph{AAAI}.

\bibitem[{Gan and Wang(2019)}]{gan2019air}
Gan, J.; and Wang, W. 2019.
\newblock In-air handwritten English word recognition using attention recurrent
  translator.
\newblock \emph{Neural Computing and Applications}, 31(7): 3155--3172.

\bibitem[{Garcia-Hernando et~al.(2018)Garcia-Hernando, Yuan, Baek, and
  Kim}]{garcia2018first}
Garcia-Hernando, G.; Yuan, S.; Baek, S.; and Kim, T.-K. 2018.
\newblock First-person hand action benchmark with rgb-d videos and 3d hand pose
  annotations.
\newblock In \emph{IEEE CVPR}, 409--419.

\bibitem[{Ge et~al.(2019)Ge, Ren, Li, Xue, Wang, Cai, and Yuan}]{ge20193d}
Ge, L.; Ren, Z.; Li, Y.; Xue, Z.; Wang, Y.; Cai, J.; and Yuan, J. 2019.
\newblock 3d hand shape and pose estimation from a single rgb image.
\newblock In \emph{IEEE CVPR}, 10833--10842.

\bibitem[{Hasson et~al.(2019)Hasson, Varol, Tzionas, Kalevatykh, Black, Laptev,
  and Schmid}]{hasson2019learning}
Hasson, Y.; Varol, G.; Tzionas, D.; Kalevatykh, I.; Black, M.~J.; Laptev, I.;
  and Schmid, C. 2019.
\newblock Learning joint reconstruction of hands and manipulated objects.
\newblock In \emph{IEEE CVPR}, 11807--11816.

\bibitem[{Hossain and Little(2018)}]{hossain2018exploiting}
Hossain, M. R.~I.; and Little, J.~J. 2018.
\newblock Exploiting temporal information for 3d human pose estimation.
\newblock In \emph{ECCV}, 68--84.

\bibitem[{Hu, Shen, and Sun(2018)}]{hu2018squeeze}
Hu, J.; Shen, L.; and Sun, G. 2018.
\newblock Squeeze-and-excitation networks.
\newblock In \emph{IEEE CVPR}, 7132--7141.

\bibitem[{Ionescu et~al.(2013)Ionescu, Papava, Olaru, and
  Sminchisescu}]{ionescu2013human3}
Ionescu, C.; Papava, D.; Olaru, V.; and Sminchisescu, C. 2013.
\newblock Human3. 6m: Large scale datasets and predictive methods for 3d human
  sensing in natural environments.
\newblock \emph{IEEE TPAMI}, 36(7): 1325--1339.

\bibitem[{Jiang et~al.(2021)Jiang, Sun, Wang, Bai, Li, and
  Fu}]{jiang2021skeleton}
Jiang, S.; Sun, B.; Wang, L.; Bai, Y.; Li, K.; and Fu, Y. 2021.
\newblock Skeleton aware multi-modal sign language recognition.
\newblock In \emph{IEEE CVPR}, 3413--3423.

\bibitem[{Kingma and Ba(2014)}]{kingma2014adam}
Kingma, D.~P.; and Ba, J. 2014.
\newblock Adam: A method for stochastic optimization.
\newblock \emph{arXiv preprint arXiv:1412.6980}.

\bibitem[{Kipf and Welling(2016)}]{kipf2016semi}
Kipf, T.~N.; and Welling, M. 2016.
\newblock Semi-supervised classification with graph convolutional networks.
\newblock \emph{arXiv preprint arXiv:1609.02907}.

\bibitem[{Li et~al.(2019)Li, Chen, Chen, Zhang, Wang, and
  Tian}]{li2019actional}
Li, M.; Chen, S.; Chen, X.; Zhang, Y.; Wang, Y.; and Tian, Q. 2019.
\newblock Actional-structural graph convolutional networks for skeleton-based
  action recognition.
\newblock In \emph{IEEE CVPR}, 3595--3603.

\bibitem[{Li, Gao, and Sang(2021)}]{li2021exploiting}
Li, M.; Gao, Y.; and Sang, N. 2021.
\newblock Exploiting Learnable Joint Groups for Hand Pose Estimation.
\newblock In \emph{AAAI}.

\bibitem[{Lin, Wang, and Liu(2021)}]{lin2021end}
Lin, K.; Wang, L.; and Liu, Z. 2021.
\newblock End-to-end human pose and mesh reconstruction with transformers.
\newblock In \emph{IEEE CVPR}, 1954--1963.

\bibitem[{Liu et~al.(2020)Liu, Ding, Zou, Wang, and
  Tang}]{liu2020comprehensive}
Liu, K.; Ding, R.; Zou, Z.; Wang, L.; and Tang, W. 2020.
\newblock A comprehensive study of weight sharing in graph networks for 3d
  human pose estimation.
\newblock In \emph{ECCV}, 318--334.

\bibitem[{Lu and Luo(2020)}]{lu2020fmkit}
Lu, D.; and Luo, L. 2020.
\newblock FMKit: An In-Air-Handwriting Analysis Library and Data Repository.
\newblock In \emph{CVPR Workshop, 2020}.

\bibitem[{Martinez et~al.(2017)Martinez, Hossain, Romero, and
  Little}]{martinez2017simple}
Martinez, J.; Hossain, R.; Romero, J.; and Little, J.~J. 2017.
\newblock A simple yet effective baseline for 3d human pose estimation.
\newblock In \emph{IEEE ICCV}, 2640--2649.

\bibitem[{Mehta et~al.(2017)Mehta, Rhodin, Casas, Fua, Sotnychenko, Xu, and
  Theobalt}]{mehta2017monocular}
Mehta, D.; Rhodin, H.; Casas, D.; Fua, P.; Sotnychenko, O.; Xu, W.; and
  Theobalt, C. 2017.
\newblock Monocular 3d human pose estimation in the wild using improved cnn
  supervision.
\newblock In \emph{IEEE 3DV}, 506--516.

\bibitem[{Mueller et~al.(2018)Mueller, Bernard, Sotnychenko, Mehta, Sridhar,
  Casas, and Theobalt}]{mueller2018ganerated}
Mueller, F.; Bernard, F.; Sotnychenko, O.; Mehta, D.; Sridhar, S.; Casas, D.;
  and Theobalt, C. 2018.
\newblock Ganerated hands for real-time 3d hand tracking from monocular rgb.
\newblock In \emph{IEEE CVPR}, 49--59.

\bibitem[{Newell, Yang, and Deng(2016)}]{newell2016stacked}
Newell, A.; Yang, K.; and Deng, J. 2016.
\newblock Stacked hourglass networks for human pose estimation.
\newblock In \emph{ECCV}, 483--499.

\bibitem[{Pavlakos et~al.(2017)Pavlakos, Zhou, Derpanis, and
  Daniilidis}]{pavlakos2017coarse}
Pavlakos, G.; Zhou, X.; Derpanis, K.~G.; and Daniilidis, K. 2017.
\newblock Coarse-to-fine volumetric prediction for single-image 3D human pose.
\newblock In \emph{IEEE CVPR}, 7025--7034.

\bibitem[{Redmon and Farhadi(2017)}]{redmon2017yolo9000}
Redmon, J.; and Farhadi, A. 2017.
\newblock YOLO9000: better, faster, stronger.
\newblock In \emph{IEEE CVPR}, 7263--7271.

\bibitem[{Romero, Tzionas, and Black(2017)}]{romero2017embodied}
Romero, J.; Tzionas, D.; and Black, M.~J. 2017.
\newblock Embodied hands: Modeling and capturing hands and bodies together.
\newblock \emph{ACM Transactions on Graphics}, 36(6): 1--17.

\bibitem[{Simon et~al.(2017)Simon, Joo, Matthews, and Sheikh}]{simon2017hand}
Simon, T.; Joo, H.; Matthews, I.; and Sheikh, Y. 2017.
\newblock Hand keypoint detection in single images using multiview
  bootstrapping.
\newblock In \emph{IEEE CVPR}, 1145--1153.

\bibitem[{Srivastava et~al.(2014)Srivastava, Hinton, Krizhevsky, Sutskever, and
  Salakhutdinov}]{srivastava2014dropout}
Srivastava, N.; Hinton, G.; Krizhevsky, A.; Sutskever, I.; and Salakhutdinov,
  R. 2014.
\newblock Dropout: a simple way to prevent neural networks from overfitting.
\newblock \emph{The journal of machine learning research}, 15(1): 1929--1958.

\bibitem[{Sun et~al.(2017)Sun, Shang, Liang, and Wei}]{sun2017compositional}
Sun, X.; Shang, J.; Liang, S.; and Wei, Y. 2017.
\newblock Compositional human pose regression.
\newblock In \emph{IEEE ICCV}, 2602--2611.

\bibitem[{Tekin, Bogo, and Pollefeys(2019)}]{tekin2019h+}
Tekin, B.; Bogo, F.; and Pollefeys, M. 2019.
\newblock H+ o: Unified egocentric recognition of 3d hand-object poses and
  interactions.
\newblock In \emph{IEEE CVPR}, 4511--4520.

\bibitem[{Tekin et~al.(2016)Tekin, Rozantsev, Lepetit, and
  Fua}]{tekin2016direct}
Tekin, B.; Rozantsev, A.; Lepetit, V.; and Fua, P. 2016.
\newblock Direct prediction of 3d body poses from motion compensated sequences.
\newblock In \emph{IEEE CVPR}, 991--1000.

\bibitem[{Vaswani et~al.(2017)Vaswani, Shazeer, Parmar, Uszkoreit, Jones,
  Gomez, Kaiser, and Polosukhin}]{vaswani2017attention}
Vaswani, A.; Shazeer, N.; Parmar, N.; Uszkoreit, J.; Jones, L.; Gomez, A.~N.;
  Kaiser, {\L}.; and Polosukhin, I. 2017.
\newblock Attention is all you need.
\newblock In \emph{Advances in neural information processing systems},
  5998--6008.

\bibitem[{Veli{\v{c}}kovi{\'c} et~al.(2017)Veli{\v{c}}kovi{\'c}, Cucurull,
  Casanova, Romero, Lio, and Bengio}]{velivckovic2017graph}
Veli{\v{c}}kovi{\'c}, P.; Cucurull, G.; Casanova, A.; Romero, A.; Lio, P.; and
  Bengio, Y. 2017.
\newblock Graph attention networks.
\newblock \emph{arXiv preprint arXiv:1710.10903}.

\bibitem[{Wang et~al.(2018)Wang, Girshick, Gupta, and He}]{wang2018non}
Wang, X.; Girshick, R.; Gupta, A.; and He, K. 2018.
\newblock Non-local neural networks.
\newblock In \emph{IEEE CVPR}, 7794--7803.

\bibitem[{Weng, Weng, and Yuan(2017)}]{weng2017spatio}
Weng, J.; Weng, C.; and Yuan, J. 2017.
\newblock Spatio-temporal naive-bayes nearest-neighbor (st-nbnn) for
  skeleton-based action recognition.
\newblock In \emph{IEEE CVPR}, 4171--4180.

\bibitem[{Xu and Takano(2021)}]{xu2021graph}
Xu, T.; and Takano, W. 2021.
\newblock Graph Stacked Hourglass Networks for 3D Human Pose Estimation.
\newblock In \emph{IEEE CVPR}, 16105--16114.

\bibitem[{Yan, Xiong, and Lin(2018)}]{yan2018spatial}
Yan, S.; Xiong, Y.; and Lin, D. 2018.
\newblock Spatial temporal graph convolutional networks for skeleton-based
  action recognition.
\newblock In \emph{AAAI}.

\bibitem[{Yang et~al.(2018)Yang, Lu, Lee, Batra, and Parikh}]{yang2018graph}
Yang, J.; Lu, J.; Lee, S.; Batra, D.; and Parikh, D. 2018.
\newblock Graph r-cnn for scene graph generation.
\newblock In \emph{ECCV}, 670--685.

\bibitem[{Zhao et~al.(2019)Zhao, Peng, Tian, Kapadia, and
  Metaxas}]{zhao2019semantic}
Zhao, L.; Peng, X.; Tian, Y.; Kapadia, M.; and Metaxas, D.~N. 2019.
\newblock Semantic graph convolutional networks for 3d human pose regression.
\newblock In \emph{IEEE CVPR}, 3425--3435.

\bibitem[{Zhou et~al.(2017)Zhou, Huang, Sun, Xue, and Wei}]{zhou2017towards}
Zhou, X.; Huang, Q.; Sun, X.; Xue, X.; and Wei, Y. 2017.
\newblock Towards 3d human pose estimation in the wild: a weakly-supervised
  approach.
\newblock In \emph{IEEE ICCV}, 398--407.

\bibitem[{Zhou et~al.(2016)Zhou, Sun, Zhang, Liang, and Wei}]{zhou2016deep}
Zhou, X.; Sun, X.; Zhang, W.; Liang, S.; and Wei, Y. 2016.
\newblock Deep kinematic pose regression.
\newblock In \emph{ECCV}, 186--201.

\end{thebibliography}

\end{document}